\documentclass[runningheads]{llncs}
\usepackage[T1]{fontenc}
\usepackage{graphicx,verbatim}
\usepackage{amssymb}
\usepackage{amsmath}
\usepackage{upgreek}
\usepackage{booktabs}
\usepackage{multirow}
\usepackage{graphicx}
\usepackage{xcolor}

\usepackage{array}
\usepackage{ulem}   
\usepackage{rotating}
\usepackage{float}
\usepackage{subcaption}
\usepackage{wrapfig}
\usepackage{booktabs}

\makeatletter
\newcommand{\equalcontrib}{\textsuperscript{*}}
\makeatother
\begin{document}
\title{\texttt{L}earning To F\texttt{ocus}: Anatomy-Guided Attention Regularization for Medical Image Classification}
%
%
\author{Tonmoy Hossain\inst{1} \and
Atiqur Rahman\equalcontrib\inst{2} \and
Farhana Hossain Swarnali\equalcontrib\inst{3} \and
Miaomiao Zhang\inst{1,4}}
%

\authorrunning{Hossain et al.}
%
\institute{
Department of Computer Science, University of Virginia, Virginia, USA \and
Ahsanullah University of Science and Technology, Dhaka, Bangladesh \and
Kahlert School of Computing, University of Utah, Utah, USA \and
Department of Electrical and Computer Engineering, University of Virginia, USA\\
\textsuperscript{*}Authors contributed equally.
}
\maketitle              
\begin{abstract}
Medical image classification models are ideally expected to identify diagnostically relevant regions while making predictions, yet standard classification losses rarely provide spatial guidance. Explicit supervision via anatomical shape information, such as segmentation masks of task-relevant anatomy, has been shown to guide the network toward regions relevant to the target prediction. However, obtaining such masks incurs substantial manual annotation effort and computational overhead. With the advent of segmentation foundation models that exhibit strong localization of anatomical structures across diverse imaging modalities, we leverage this capability to extract anatomical shape priors without the burden of training a dedicated segmentation model. In this paper, we propose a new framework \texttt{Locus}, \textit{an anatomical attention regularization approach that leverages pretrained segmentation foundation models to guide a classifier's attention toward diagnostically meaningful anatomical structures across diverse imaging modalities}. Instead of enforcing pixel-wise alignment with the predicted anatomical mask, we introduce a regularization term that adaptively balances network's attention between anatomical (foreground) and background regions, penalizing the classifier when background attention dominates. We validate \texttt{Locus} on eight diverse medical imaging datasets spanning dermoscopy, X-ray, histopathology, and cardiac MRI, showing consistent gains in classification performance alongside improved anatomically grounded attention. Our code is available at \texttt{https://github.com/tonmoy-hossain/Locus}.

\keywords{Image Classification \and Attention Regularization \and Anatomical Shapes \and Medical Imaging.}
\end{abstract}

\section{Introduction}

Reliable image classification is central to many high-stakes medical applications, from cancer screening to disease diagnosis, yet medical image classifiers are typically trained with only image-level labels, leaving the network to implicitly determine which regions are diagnostically relevant without any direct spatial guidance~\cite{hossain2024invariant,hossain2019brain,hossain2026learning,ling2023mtanet,muller2022joint,yang2025diffmic}. In the absence of explicit spatial supervision, models tend to learn attention patterns primarily from statistical correlations rather than anatomically meaningful cues. As a result, their attention may be distributed across anatomically irrelevant regions, such as surrounding tissue, imaging artifacts, or dataset-specific contextual features, limiting both interpretability and anatomical grounding~\cite{ahmed2021automated,li2025towards,lin2025cxr,yang2025diffmic}. 

\noindent{\bf Related works.} To address this, prior works have explored auxiliary anatomical shape information, for example, encoded as segmentations of relevant structures, to guide classifier attention toward relevant anatomy~\cite{gao2025multiscale,inan2022deep}. 
To provide the spatial guidance that existing networks lack, a line of work has investigated different ways of directly supervising the classifier's attention with anatomical structure. First, manually delineated anatomical masks directly constrain the classifier during training~\cite{ling2023mtanet,muller2022joint,noothout2020deep}, but require expert clinicians to annotate the entire training set, making them costly and difficult to scale to large datasets. To reduce this annotation cost, subsequent works instead train a segmentation branch jointly with the classifier and use its predicted masks as the source of anatomical guidance~\cite{gao2025multiscale,inan2022deep}. However, jointly optimizing segmentation and classification is difficult, as the two losses compete for shared gradients and require careful weighting, while noisy segmentation predictions can corrupt the anatomical guidance passed to the classifier~\cite{gao2025multiscale,inan2022deep,mahbod2020effects}. Segmentation foundation models (SFM), such as SAM and their medical-domain adaptations~\cite{huang2024segment,jiang2026medical,kirillov2023segment,ma2024segment,zhao2025foundation}, have since revolutionized anatomical localization, offering strong, out-of-the-box segmentation across imaging modalities without any dedicated training, making them a reliable and lightweight source of anatomical guidance for the classifier.
\begin{figure}[!b]
    \centering
    \includegraphics[width=1.0\linewidth, trim=0 0 0 0cm]{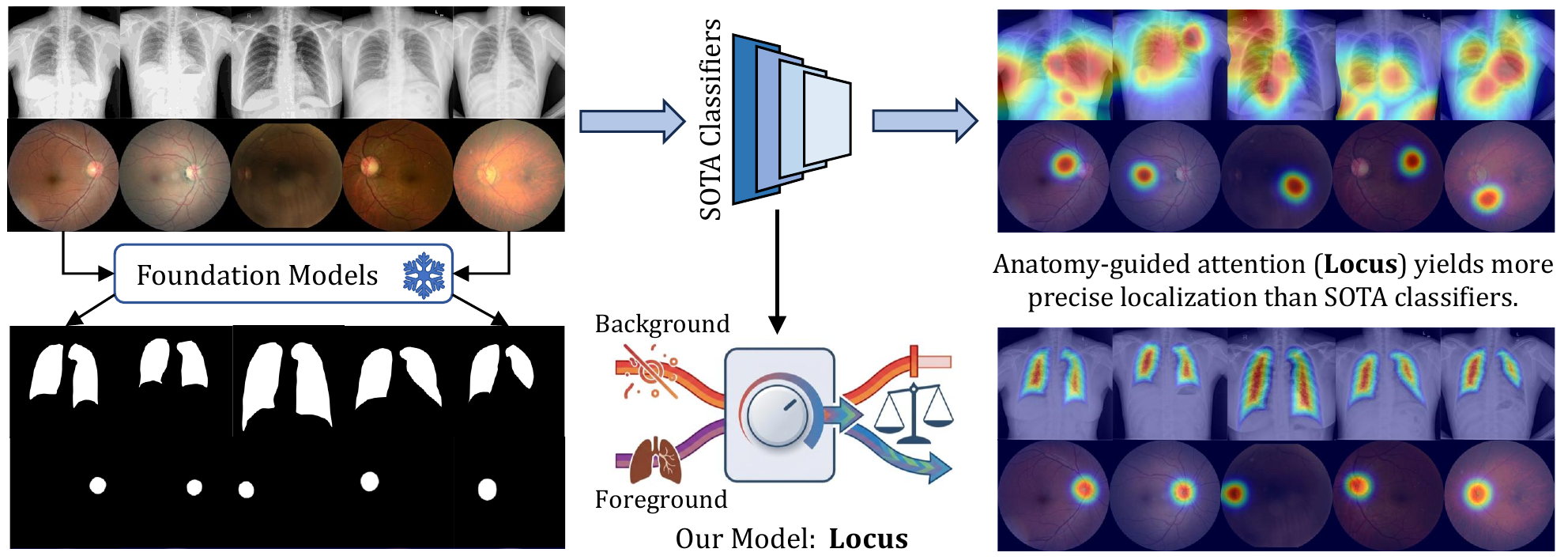}
    \caption{Visualization of overlaid attention maps from SOTA classifiers vs. our anatomy-guided model (\texttt{Locus}), which reweights foundation-model features toward the anatomical regions (e.g., lung and optic disc/cup) for improved performance with better interpretability.}
    \label{fig:intFig}
\end{figure}

Another line of research leveraged Class Activation Mapping (CAM)-based pseudo-masks to constrain focus without requiring pixel-level annotations~\cite{jiang2020multi,selvaraju2020grad}. However, these approaches face a fundamental trade-off between annotation cost and spatial coverage: auxiliary segmentation branches achieve near-perfect anatomical coverage but require expensive pixel-level annotations for training, whereas CAM-based pseudo-masks avoid this annotation cost but tend to activate only the discriminative regions correlated with the target label, rather than the anatomical structures relevant to the prediction ~\cite{madan2025focus,xie2024rethinking,zhang20213d}. Alternatively, one could use manually annotated anatomical masks, as in the recent MGA~\cite{uddin2025expert}, or obtain them from a dedicated segmentation model~\cite{patricio2023explainable}. However, both approaches face practical constraints: the former requires extensive manual effort, while the other requires additional model training and computational resources.

Harnessing the strong anatomical localization capabilities of SFMs, we introduce \texttt{Locus}, that for the first time regularizes the classifier's attention toward class-relevant anatomical shape structures, entirely removing the costly requirement of dedicated segmentation training (as shown in Fig.~\ref{fig:intFig}). Specifically, \texttt{Locus} introduces a regularization term that adaptively balances attention between anatomical (foreground) and background regions, avoiding rigid pixel-wise alignment with the foundation-model-derived mask and instead penalizing the classifier only when background attention dominates. Our contributions are three-fold:
\begin{itemize}
    \item Develop an anatomy-guided attention regularization framework, \texttt{Locus}, that adaptively
    balances classifier attention between foreground and background regions using SFM
    outputs, penalizing the classifier only when background attention dominates.
    \item Easily adaptable to a variety of classifier backbones and SFMs,
    serving as a plug-and-play regularization module that requires no modification to the
    classifier or fine-tuning of the SFM to boost classification performance.
    \item Validate on eight diverse medical imaging datasets spanning X-ray, histopathology, and cardiac MRI, demonstrating consistent gains in classification performance and more anatomically grounded, interpretable attention.
\end{itemize}

\section{Our Method}
In this section, we present \texttt{Locus}, a new framework to guide medical image classifiers toward diagnostically relevant anatomical regions using anatomical priors derived from pre-trained segmentation foundation models. We organize this section as (i) developing the anatomy-guided attention regularization term, (ii) describing our patch-wise anatomical prompting strategy for extracting anatomical masks from segmentation foundation models, and (iii) designing the overall network loss.\\

\noindent{\textbf{Problem Formulation.}} Let $\mathbf{x} \in \mathbb{R}^{H \times W \times D}$ denote an input image, where $D$ represents the number of channels, temporal frames, or volumetric slices depending on the modality, and $y \in \{1, \ldots, C\}$ its class label in a $C$-class classification task. A classification network with encoder $\mathcal{E}$, parameterized by $\Uptheta_{\mathcal{E}}$, produces predictions $\hat{y} = f(\mathbf{x};{\Uptheta}_\mathcal{E})$ via a forward pass. During training, $\Uptheta_{\mathcal{E}}$ is optimized by minimizing the standard cross-entropy loss
\begin{equation}
    \mathcal{L}_{\text{task}}(\mathbf{x}; \Uptheta_{\mathcal{E}}) = -\sum_{c=1}^{C} \mathbb{I}_{[y=c]} \log p_c(\mathbf{x}; \Uptheta_{\mathcal{E}}),
\end{equation}
where $p_c(\mathbf{x}; \Uptheta_{\mathcal{E}})$ is the predicted probability for class $c$. While minimizing this empirical loss may encourage accurate predictions, it provides no spatial supervision, leaving the network free to attend to anatomically irrelevant regions. To address this, let $\mathbf{A}(\mathbf{x}; \Uptheta_{\mathcal{E}})$ denote the spatial attention map derived from $\mathcal{E}$ and $\mathbf{M} = \mathcal{S}(\mathbf{x})$ the anatomical mask from a pre-trained, frozen segmentation foundation model $\mathcal{S}$, with the objective of encouraging attention within $\mathbf{M}$ while suppressing it in background regions. The details of our network architecture are introduced below in Fig.~\ref{fig:model}.
\begin{figure}[!b]
    \centering
    \includegraphics[width=\linewidth, trim=0 0 0 0]{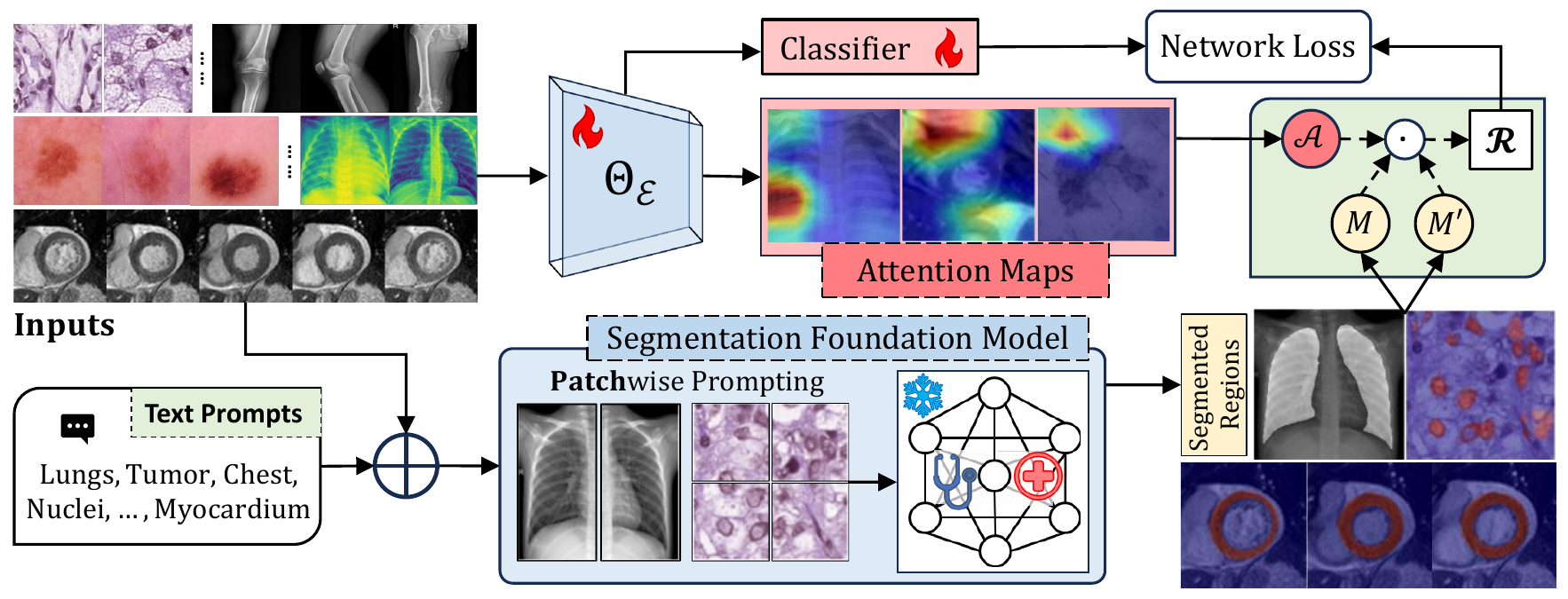}
    \caption{An overview of the proposed \texttt{Locus}. The image encoder $\mathcal{E}$ produces class predictions and an attention map $\mathcal{A}$, while a segmentation foundation model, guided by patch-wise text prompts (e.g., lungs, tumor, ..., myocardium), produces the anatomical foreground mask $M$ and its complement background mask $M' = 1-M$. The anatomy-guided attention regularizer $\mathcal{R}$ penalizes the network when on $M$ is biased by $M'$.}
    \label{fig:model}
\end{figure}

\subsection{Anatomy-guided Attention Regularization}
\label{sec:anatomy_guided_att_reg}
Since a classifier trained with the standard ERM (Empirical Risk Minimization) objective is optimized only with an image-level cross-entropy loss, its attention naturally concentrates on regions that are statistically correlated with the target labels, irrespective of their anatomical validity~\cite{degrave2021ai,geirhos2020shortcut,zech2018variable}. However, many classification tasks are inherently anatomy-specific: distinguishing pneumonia from healthy patients depends on the lung regions, grading a skin lesion depends on the lesion boundary, and identifying a bone tumor depends on the affected bone region~\cite{jiang2024anatomy,muller2023anatomy,rajpurkar2018deep}. Aligning network attention with such diagnostically relevant anatomical structures is therefore necessary to prevent the classifier from relying on unreliable spurious features, such as imaging artifacts or scanner-specific cues, that fail to generalize across scanners, sites, or patient populations. This motivates obtaining an attention map to compare directly against these diagnostically relevant anatomies.

Recall from our problem formulation that $\mathbf{A}(\mathbf{x}; \Uptheta_{\mathcal{E}})$ denotes the spatial attention map derived from the encoder $\mathcal{E}$, and $\mathbf{M} = \mathcal{S}(\mathbf{x})$ the anatomical mask from the frozen segmentation foundation model $\mathcal{S}$. We instantiate $\mathbf{A}(\mathbf{x}; \Uptheta_{\mathcal{E}})$ using a class-discriminative saliency method: $\mathbf{A}(\mathbf{x}; \Uptheta_{\mathcal{E}}) = \text{Attn}(\mathcal{E}(\mathbf{x}; \Uptheta_{\mathcal{E}}))$. We now quantify alignment between attention and the anatomical mask $\mathbf{M} \in \{0, 1\}^{H \times W}$, obtained from the segmentation foundation model, by computing weighted averages over foreground and background regions as

\begin{equation}
\mathcal{A}_{fg}(\mathbf{x}) = \frac{\sum_i \mathbf{M}_i \cdot \mathbf{A}_i(\mathbf{x})}{\sum_i \mathbf{M}_i + \epsilon}, \quad
\mathcal{A}_{bg}(\mathbf{x}) = \frac{\sum_i (1 - \mathbf{M}_i) \cdot \mathbf{A}_i(\mathbf{x})}{\sum_i (1 - \mathbf{M}_i) + \epsilon},
\label{eq:fg_bg}
\end{equation}
where $i$ indexes spatial locations, $\mathbf{A}_i(\mathbf{x})$ denotes the value of $\mathbf{A}(\mathbf{x}; \Uptheta_{\mathcal{E}})$ at location $i$, and $\epsilon$ is a small constant for numerical stability. Rather than enforcing pixel-wise alignment, this formulation compares aggregate attention energy within and outside the anatomical region, making the supervision robust to imprecise mask boundaries and spatially varying attention distributions.

In contrast to conventional approaches that apply uniform pixel-wise spatial constraints, we define the anatomy-guided regularization term as
\begin{equation}
\mathcal{R}(\mathbf{A}(\mathbf{x})) = \max(0, \mathcal{A}_{bg}(\mathbf{x}) - \mathcal{A}_{fg}(\mathbf{x})),
\label{eq:regularizer}
\end{equation}
where $\mathcal{A}_{fg}(\mathbf{x})$ and $\mathcal{A}_{bg}(\mathbf{x})$ denote the classifier's attention mass over the foreground and background regions of $\mathbf{x}$, as defined by the predicted anatomical mask. This hinge formulation remains inactive whenever $\mathcal{A}_{fg} \geq \mathcal{A}_{bg}$, avoiding over-suppression of contextual information that may be diagnostically relevant, and only activates a corrective gradient when background attention dominates. We adopt this soft formulation by design, so that $\mathcal{R}$ corrects only the specific failure mode where background attention dominates, rather than driving all attention onto the anatomical region. In contrast to pixel-wise alignment objectives (e.g., Dice loss), which penalize any deviation from the mask regardless of diagnostic relevance, $\mathcal{R}$ requires only that attention remain anatomy-dominant rather than anatomy-exclusive, without introducing any additional learnable parameters. Importantly, $M = \mathcal{S}(\mathbf{x}; \Theta_\mathcal{S})$ is produced by a frozen segmentation foundation model that is never optimized jointly with the classifier. Consequently, $M$ never modifies $\mathbf{x}$ (e.g., via masking or as an auxiliary input channel); it is used only to compute $A_{fg}$ and $A_{bg}$ (Eq.~\eqref{eq:fg_bg}), which define $\mathcal{R}$ (Eq.~\eqref{eq:regularizer}). As a result, gradients from $\mathcal{R}$ propagate only to $\Theta_{\mathcal{E}}$. We apply a stop-gradient to the Grad-CAM computation, so that $\mathcal{R}$ backpropagates only through network $\mathcal{E}$, not through Grad-CAM's gradient-based channel weights.

\subsection{Patch-wise Anatomical Prompting} 
\label{sec:PAP}
\begin{figure}[h]
    \centering
    \includegraphics[width=\linewidth, trim=0 0 0 1cm]{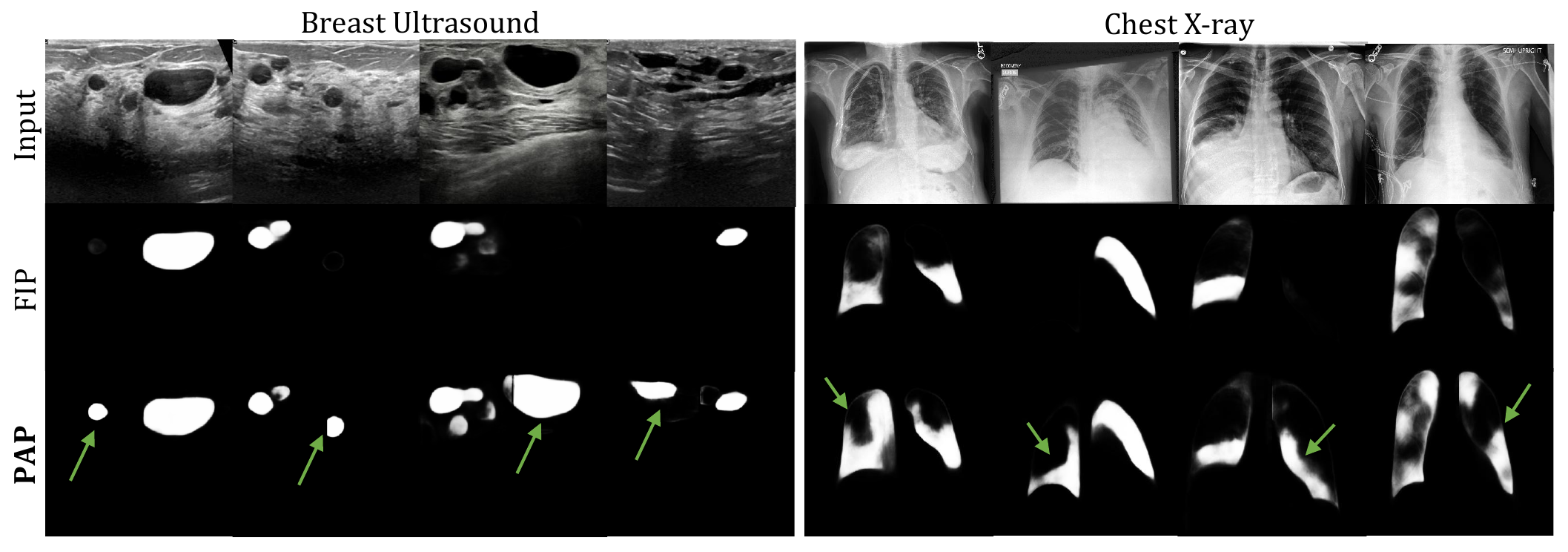}
    \caption{Comparison of anatomical masks obtained from the pretrained BiomedParse~\cite{zhao2025foundation} model via Full-Image Prompting (FIP) and our Patch-wise Anatomical Prompting (PAP) on Breast Ultrasound and Chest X-ray images, with arrows highlighting anatomical regions recovered by PAP but missed by FIP.}
    \label{fig:patchPrompt}
\end{figure}

Segmentation foundation models such as Medical SAM3~\cite{jiang2026medical} or BiomedParse~\cite{zhao2025foundation}, when prompted on the full image, often fail to capture all relevant anatomical structures (Fig.~\ref{fig:patchPrompt}), either returning a single dominant region or producing degenerate masks when the target anatomy is small relative to the image canvas. To address this without fine-tuning or architectural modification, we propose a simple yet effective \textit{patch-wise anatomical prompting} strategy. Given an input image $\mathbf{x}$, we partition it into a uniform grid of $P$ non-overlapping patches $\{x_p\}_{p=1}^{P}$, with $P$ chosen empirically per dataset based on the expected scale of the target anatomy. Each patch is independently prompted with the same class-derived text descriptor (e.g., ``tumor lesion'', ``lungs'', etc.), and the resulting patch-level masks are aggregated via union to form the full-resolution object mask $\mathbf{M} = \bigcup_{p=1}^{P} \mathcal{S}(x_p; \Uptheta_S)$, union aggregation maximizes anatomical coverage, ensuring all relevant regions contribute to attention supervision regardless of spatial distribution. Crucially, \textit{our goal is not precise segmentation, but sufficient anatomical grounding to guide classifier attention}; even coarse or partial masks suffice, since the hinge formulation in $\mathcal{R}$ penalizes only substantial misallocation of attention to the background, not precise pixel-level correspondence.

As illustrated in Fig.~\ref{fig:patchPrompt}, when an image contains multiple disjoint instances of the same anatomical structure, such as scattered lesions or bilateral lung fields, prompting a pre-trained segmentation foundation model (e.g., BiomedParse~\cite{zhao2025foundation} or Medical SAM~\cite{jiang2026medical}) on the full image (FIP) often captures the most dominant instance, leaving other anatomically relevant regions unsegmented. Our patch-wise prompting (PAP) strategy addresses this by independently prompting each spatial patch, recovering the missed regions (\textcolor{green}{$\nearrow$}) and yielding more complete anatomical masks for guiding the regularizer. \textit{Please note that this prompting strategy is not regularized by any learnable parameters, and future work could explore learning the patch configuration end-to-end}; our primary motivation here is to introduce a simple yet efficient strategy for obtaining reliable anatomical guidance from a pre-trained SFM. 

We find that segmentation quality from the pre-trained SFM can remain coarse or partial even under patch-wise prompting, as seen for chest X-ray lung masks in Fig.~\ref{fig:patchPrompt}, though such coarseness is tolerable since $\mathcal{R}$ only penalizes substantial misalignment of attention with the background rather than requiring precise pixel-level correspondence (Sec.~\ref{sec:anatomy_guided_att_reg}).
  
\subsection{Network Loss}

Now we are ready to define our network loss by combining the classification loss with the anatomy-guided regularizer,
\begin{equation}
\mathcal{L}_{\text{net}}(\Theta_{\mathcal{E}}) = \mathcal{L}_{\text{task}}(\mathbf{x}; \Theta_{\mathcal{E}}) + \lambda \cdot \mathcal{R}(\mathbf{A}(\mathbf{x})).
\label{eq:net_loss}
\end{equation}
where $\lambda$ controls the regularization strength. We employ a warm-up strategy where the network is first trained with $\mathcal{L}_{\text{task}}$ alone for $T_{\text{warmup}}$ epochs, then the regularization is gradually increased to $\lambda_{\max}$ over $T_{\text{ramp}}$ epochs. Object masks $\mathbf{M}$ are pre-computed offline, avoiding any additional computational overhead during training. While we primarily use Grad-CAM~\cite{selvaraju2017grad} to derive $\mathbf{A}$, our framework is agnostic to the choice of attention extraction method, and other saliency approaches such as GradCAM++~\cite{chattopadhay2018grad}, Vanilla CAM~\cite{zhou2016learning}, or Gradient-Norm can be seamlessly integrated without any modification to $\mathcal{R}$.

\section{Experimental Evaluation}
\noindent{\textbf{Datasets.}} We evaluate our model across a comprehensive suite of public and private medical imaging datasets covering diverse distinct imaging modalities, anatomical structures, and both \textbf{static image} and \textbf{dynamic video} inputs, providing far stronger evidence of generalizability than single-domain evaluations. The seven public datasets span (i) \textbf{HAM10000}~\cite{tschandl2018ham10000} ($10015$ dermoscopic images, $7$-class skin lesion classification: Actinic Keratoses, Basal cell carcinoma, Benign keratosis, Dermatofibroma, Melanocytic nevi, Melanoma, and Vascular skin); (ii) \textbf{PneumoniaMNIST}~\cite{kermany2018identifying,yang2023medmnist} ($5856$ pediatric chest X-rays, binary pneumonia classification against normal); (iii) \textbf{ChestX-ray8}~\cite{wang2017chestx} ($112{,}120$ frontal-view chest X-rays from NIH-ChestXray14, multi-label classification of $14$ thoracic pathologies);  (iv) \textbf{CheXpert}~\cite{irvin2019chexpert} ($1689$ chest X-rays, multi-label classification over $4$ observations: edema, atelectasis, pleural effusion and cardiomegaly); (v) \textbf{BTXRD}~\cite{yao2025radiograph} ($3746$ multi-center musculoskeletal X-rays, binary bone tumor classification: benign or malignant); (vi) \textbf{PanNuke}~\cite{gamper2019pannuke} ($7904$ H\&E-stained histopathology patches, nuclear type classification across $19$ tissue types and $5$ cell categories, including neoplastic/inflammatory/connective/dead/epithelial); (vii) \textbf{BUSI}~\cite{al2020dataset} ($780$ breast ultrasound images, two-class breast lesion classification: benign and malignant).

We additionally evaluate on an \textbf{in-house cardiac cine MRI dataset}~\cite{wang2022ai} comprising $510$ video sequences, having $24$ timeframes per video, for binary classification of myocardial scar presence versus healthy myocardium. Unlike all other datasets, this task operates on temporal image sequences, requiring the model to capture both spatial morphology and cardiac motion dynamics, demonstrating the applicability of our approach to time-series medical data. 

Please note that we excluded images whose class labels are not primarily determined by anatomical shape or lesion morphology. For example, we exclude the `No Finding class' in ChestX-ray8 and the `normal class' in BUSI, as they do not correspond to a specific anatomical abnormality or lesion when performing diagnostic classification.

\subsection{Experimental Settings}
\noindent{\textbf{Classification Performance.}}
We evaluate the effectiveness of our framework by performing classification across all datasets under three CNN backbones: ResNet~\cite{he2016deep}, EfficientNet~\cite{tan2019efficientnet}, and DenseNet~\cite{huang2017densely}, reporting accuracy and F1-score. We compare against three baselines, including \textbf{CE}: optimizes the cross-entropy loss solely without any spatial supervision; \textbf{Masked}: multiplies the input image by the SFM-derived mask (setting background pixels to zero) before passing it to the classifier, considering only anatomical regions and discarding potentially relevant contextual information; and \textbf{MGA}~\cite{uddin2025expert}: utilizes a Grad-CAM-based Dice loss to align attention with radiologist-annotated ROI masks, jointly optimized with a prototypical few-shot classification objective, requiring expert annotations rather than SFM-derived masks.

\noindent{\textbf{Attention Visualization and Analysis.}}
To qualitatively validate that our approach guides models toward anatomically relevant regions, we visualize and compare attention maps of our model against all baselines using GradCAM~\cite{selvaraju2017grad}.

\noindent\textbf{Effectiveness of Our Regularizer.} To isolate the contribution of our proposed regularizer $\mathcal{R}$, we compare classifiers trained with task-specific loss ($\mathcal{L}_{\text{task}}$) (for classification, a standard cross-entropy term) alone against Ours ($\mathcal{L}_{\text{task}} + \mathcal{R}$) across convolutional (ResNet, ConvNeXt~\cite{liu2022convnet}) and transformer-based (ViT~\cite{dosovitskiy2020image}) backbones. We visualize GradCAM attention maps on Breast Ultrasound and Chest X-ray images for each backbone to assess how much anatomical grounding the regularizer recovers relative to $\mathcal{L}_{\text{task}}$.

For this ablation study, we additionally consider BiomedParse~\cite{zhao2025foundation} as an alternative segmentation foundation model (SFM) to Medical SAM, further demonstrating the benefits of our regularizer generalize across different underlying SFM choices.

\noindent{\textbf{Generalizability of Attention Regularization.}}
To demonstrate that our anatomy-guided regularization is not tied to any specific saliency method, we evaluate its compatibility beyond GradCAM by presenting attention visualizations using Vanilla CAM~\cite{zhou2016learning}, Gradient Norm, and GradCAM++~\cite{chattopadhay2018grad}. For each method, we quantitatively report classification performance and qualitatively compare the resulting attention maps with and without our proposed regularization, showing that the anatomically coherent behavior induced by our approach generalizes across diverse visualization techniques.

\noindent\textbf{Ablation and Robustness Analysis.} To assess the sensitivity of our framework to the regularization weight $\lambda$, we ablate its value on BUSI dataset under ResNet backbone, reporting accuracy, precision, and F1-score against the CE-only baseline.

We further evaluate robustness under distribution shift by testing our model on Chest X-ray images corrupted with varying levels of brightness, contrast, Gaussian blur, and Gaussian noise, applied only at test time, comparing Ours against the CE-only baseline across ResNet, ConvNeXt, and ViT backbones. We also analyze failure cases of SFMs to better understand the limitations of our prompting strategies.\\

\noindent\textit{Implementation Details.} For classification, we employ standard CNN backbones with cross-entropy loss $\mathcal{L}_{\text{task}}$, optimizing parameters using the Adam optimizer~\cite{kingma2014adam}. For segmentation foundation models, we primarily use Medical SAM3~\cite{jiang2026medical} and further evaluate generalizability on BiomedParse~\cite{zhao2025foundation}. However, our framework is agnostic to the specific choice of segmentation foundation model and generalizes to any pretrained text-guided segmentation model. We apply a warm-up strategy by training with $\mathcal{L}_{\text{task}}$ alone for $T_{\text{warmup}} = 5$ epochs, then ramp regularization strength $\lambda$ from $0$ to $\lambda_{\max} = 0.1$ over $T_{\text{ramp}} = 5$ epochs. We use this same fixed $\lambda_{\max}$ across all datasets and backbones, without dataset-specific tuning on a held-out validation split, to demonstrate that \texttt{Locus} is effective under a single global hyperparameter setting; we separately study sensitivity to this choice in our ablation analysis. Note that $\lambda$ only affects training via the regularizer $\mathcal{R}$ in Eq.~\eqref{eq:net_loss} and plays no role at test time, where inference reduces to a standard forward pass through the trained classifier. We extract attention maps $\mathbf{A}$ using Grad-CAM~\cite{selvaraju2017grad} from the final convolutional layer. We train all models for 100 epochs with learning rate $10^{-4}$ and batch size $32$. We use standard train/validation/test splits provided by each dataset, or apply $70/15/15$ random splits with a fixed seed where predefined splits are unavailable.

\subsection{Results}

Tab.~\ref{tab:acc_perf} reports classification accuracy and F1-score across six benchmark datasets spanning dermoscopy, X-ray, histopathology, and cardiac MRI, comparing our method against standard baselines. We achieve the highest accuracy and F1-score in nearly all settings, outperforming unconstrained training, masked-input, and mask-guided attention baselines, demonstrating the effectiveness of the proposed approach across heterogeneous modalities.

\begin{table}[h]
\centering
\renewcommand{\arraystretch}{1.2}
\caption{Classification performance comparison across six medical imaging datasets under three CNN backbones.}
\resizebox{\textwidth}{!}{%
\begin{tabular}{lccccccccccccc}
\hline
\multirow{2}{*}{\textbf{Dataset}}
  & \multirow{2}{*}{\textbf{Metric}}
  & \multicolumn{4}{c}{\uline{ResNet}}
  & \multicolumn{4}{c}{\uline{EfficientNet}}
  & \multicolumn{4}{c}{\uline{DenseNet}} \\

 &
  & \textit{Image} & \textit{Masked} & \textit{MGA} & \textbf{Ours}
  & \textit{Image} & \textit{Masked} & \textit{MGA} & \textbf{Ours}
  & \textit{Image} & \textit{Masked} & \textit{MGA} & \textbf{Ours} \\
\hline

\multirow{2}{*}{HAM10000}
  & \textit{Acc}   & 83.99 & 76.88 & 81.15 & \textbf{86.53} & 86.27 & 84.04 & 84.21 & \textbf{89.28} & 86.06 & 85.04 & 82.44 & \textbf{88.48} \\
  & \textit{F1-sc} & 83.19 & 73.61 & 79.79 & \textbf{86.07} & 82.55 & 83.60 & 80.47 & \textbf{89.04} & 81.53 & 84.60 & 81.31 & \textbf{88.17} \\
\hline

\multirow{2}{*}{PneumoniaMNIST}
  & \textit{Acc}   & 89.90 & 90.54 & 89.90 & \textbf{92.15} & 91.15 & 91.31 & 91.47 & \textbf{92.79} & 91.03 & 91.03 & 91.03 & \textbf{92.15} \\
  & \textit{F1-sc} & 89.53 & 90.24 & 89.53 & \textbf{91.93} & 90.96 & 91.11 & 91.18 & \textbf{92.62} & 91.79 & 91.75 & 91.79 & \textbf{91.96} \\
\hline

\multirow{2}{*}{BTXRD}
  & \textit{Acc}   & 81.49 & 85.05 & 86.48 & \textbf{87.54} & 81.85 & 83.99 & 86.24 & \textbf{91.49} & 81.49 & 83.27 & 85.05 & \textbf{88.61} \\
  & \textit{F1-sc} & 83.19 & 84.27 & 83.66 & \textbf{86.24} & 74.03 & 81.09 & 85.16 & \textbf{91.10} & 76.19 & 81.30 & 83.93 & \textbf{88.51} \\
\hline

\multirow{2}{*}{PanNuke}
  & \textit{Acc}   & 76.53 & 73.98 & 76.41 & \textbf{82.41} & 81.93 & 76.87 & 81.57 & \textbf{88.80} & 87.47 & 80.24 & 87.35 & \textbf{90.12} \\
  & \textit{F1-sc} & 75.83 & 71.31 & 75.22 & \textbf{82.15} & 81.68 & 76.03 & 81.31 & \textbf{88.57} & 87.08 & 79.28 & 86.95 & \textbf{90.03} \\
\hline

\multirow{2}{*}{ChestX-ray8}
  & \textit{Acc}   &  $80.64$     & $81.52$       & $79.65$       & \textbf{88.28} & 85.71 & 85.85 &  $\textbf{88.57}$     & 87.83 &   $77.85$     & $84.80$      & $84.59$      & \textbf{86.28} \\
  & \textit{F1-sc} & $82.76$       & $83.39$       & $82.25$       & \textbf{86.87} & 86.05 & 85.23 & $\textbf{87.83}$      & 86.74 & $80.83$     & $82.51$        & $82.80$        & \textbf{84.68} \\
\hline

\multirow{2}{*}{Cardiac}
  & \textit{Acc}   & 76.40 & 79.78 & 79.78 & \textbf{83.15} & 81.69 & 81.23 & 81.65 & \textbf{84.27} & 78.32 & 80.88 & 78.37 & \textbf{82.02} \\
  & \textit{F1-sc} & 75.76 & 79.64 & 79.44 & \textbf{82.81} & 81.69 & 81.23 & 81.63 & \textbf{84.01} & 78.33 & 80.86 & 78.14 & \textbf{81.30} \\
\hline

\end{tabular}%
}%
\label{tab:acc_perf}
\end{table}

\begin{figure}[!ht]
    \centering
    \includegraphics[width=0.94\linewidth, trim=0 0 0 0]{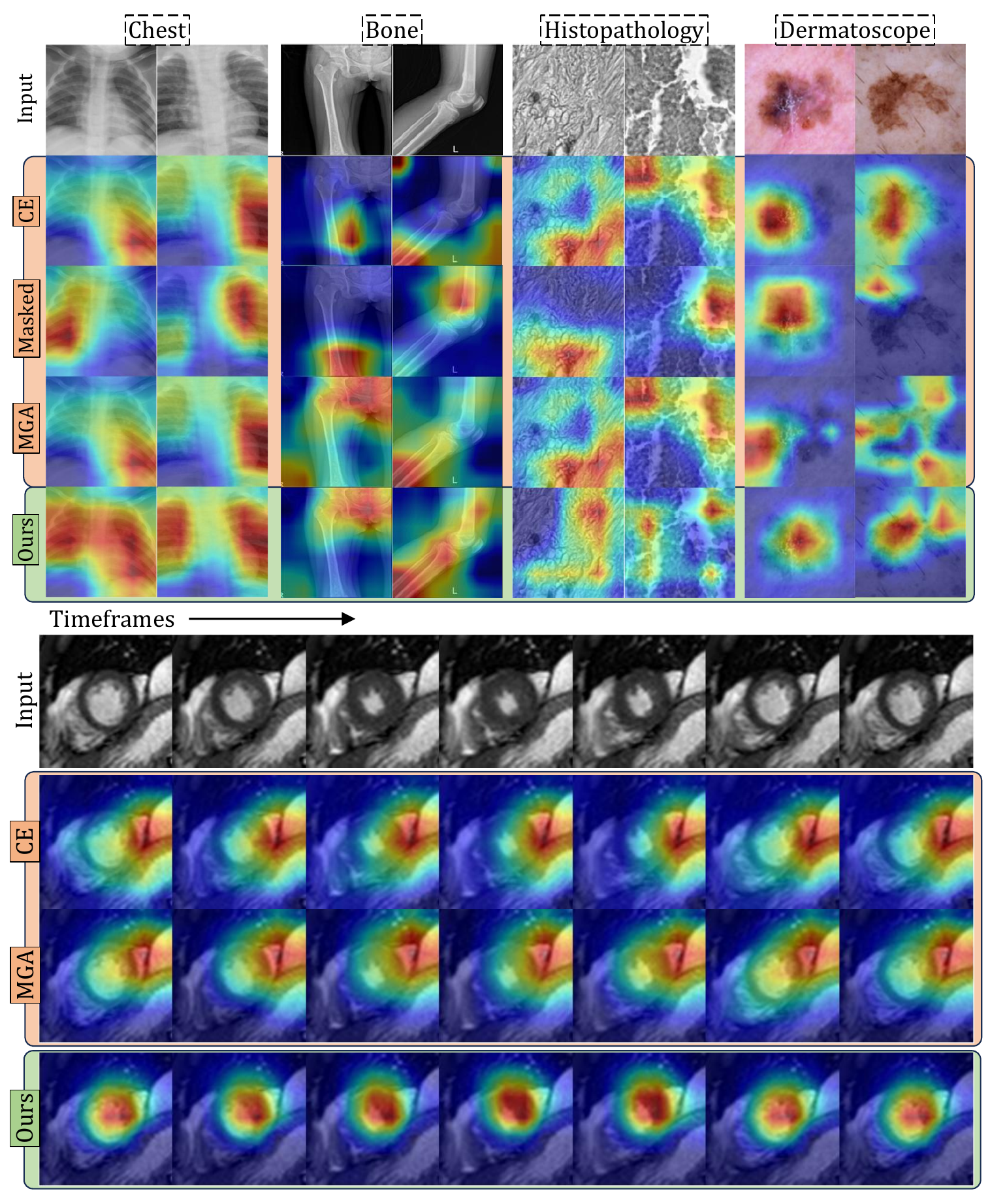}
    \caption{GradCAM visualizations across four imaging modalities and cardiac cine MRI videos, comparing CE, Masked, MGA, and Ours. Top: Chest, Bone, Histopathology, and Dermatoscope (left to right). Bottom: Cardiac cine MRI across temporal timeframes.}
    \label{fig:gradcam_viz}
\end{figure}

Fig.~\ref{fig:gradcam_viz} shows GradCAM visualizations across four imaging modalities and cardiac cine MRI timeframes, comparing CE, Masked, MGA, and Ours. \texttt{Locus} consistently produces more focused and anatomically coherent attention maps, concentrating activation on diagnostically relevant structures while suppressing unrelated background responses across most modalities and temporal frames. Notably, for cardiac cine MRI, \texttt{Locus} activation remains centered on the myocardium across all timeframes despite cardiac motion, while CE and MGA activation spreads into surrounding chambers as the cycle progresses, which is directly relevant for myocardial scar classification since the diagnostic signal is localized to the myocardial tissue itself. For dermoscopy, \texttt{Locus} attention more completely covers spatially disjoint lesion components, whereas baselines attend to only part of the lesion or activate spurious background regions; for chest X-ray, \texttt{Locus} spans both lung fields while baselines concentrate on a single lung; for bone, CE frequently attends to soft tissue away from the bone, while \texttt{Locus} attention follows the bone contour more closely. For histopathology, differences between methods are visually subtle, as the fine-grained, spatially distributed nature of nuclei offers no single dominant landmark for qualitative comparison.

Note that despite background pixels being zeroed for the Masked baseline, its Grad-CAM heatmaps can still extend beyond the anatomical region, since Grad-CAM is computed from a coarse feature map and upsampled to input resolution, and imprecision in the SFM-derived mask $M$ can leave residual background in the masked input.

\begin{figure}[!ht]
    \centering
    \includegraphics[width=0.95\linewidth, trim=0 0 0 0]{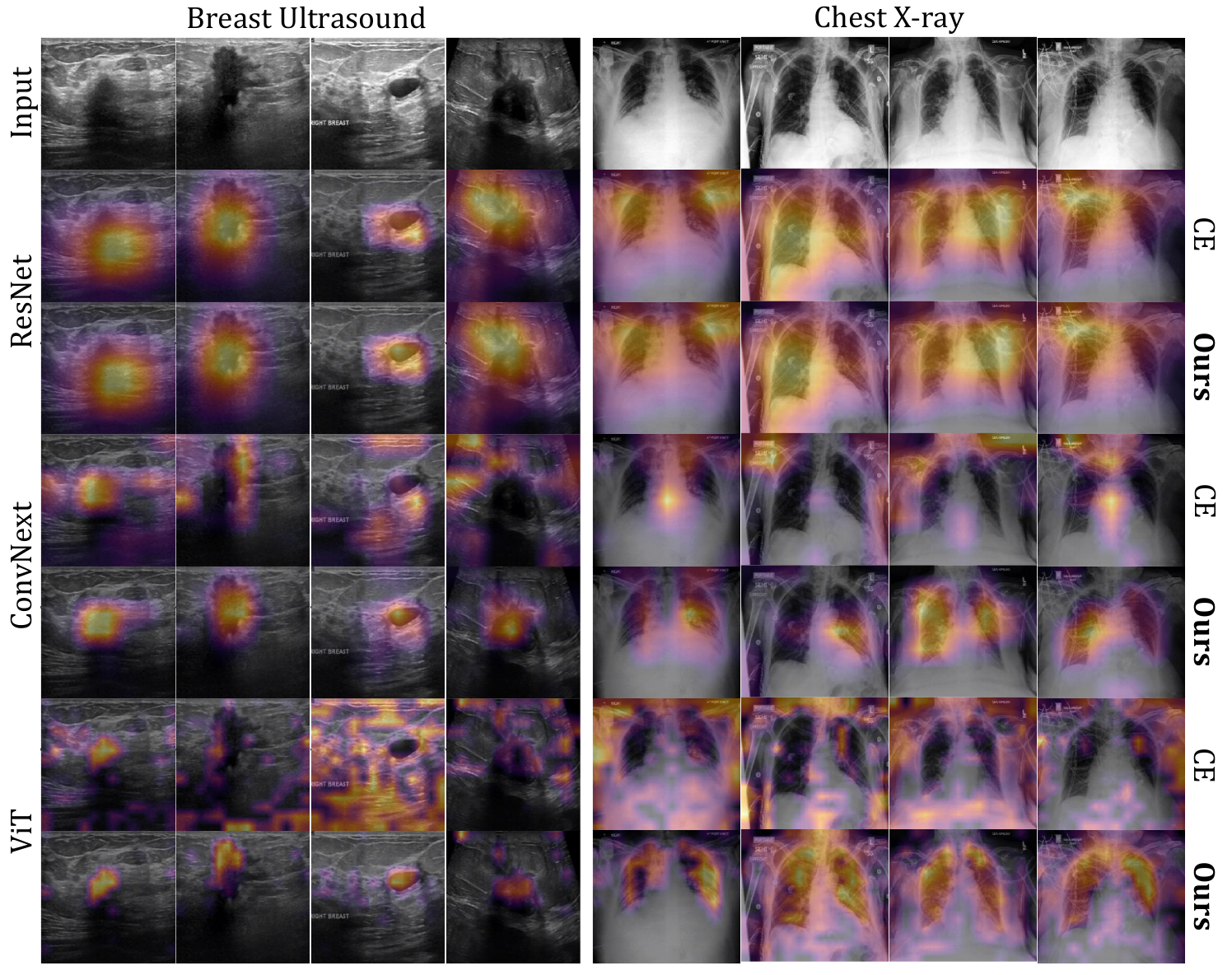}
    \caption{Comparison of GradCAM visualizations between network trained with Cross-Entropy (CE) loss and Ours on Breast Ultrasound (BUSI) and Chest X-ray images (CheXpert dataset), across three backbones.}
    \label{fig:ablgradcam}
\end{figure}

Fig.~\ref{fig:ablgradcam} examines attention quality across three backbones, including ResNet, ConvNeXt, and ViT, on Breast Ultrasound and Chest X-ray images. Under the ResNet backbone, CE attention largely matches Ours, remaining reasonably focused on the lesion and lung regions. However, for ConvNeXt  and ViT, CE attention becomes noticeably diffuse, spreading onto shadowing artifacts and soft tissue in ultrasound and beyond the lung fields in X-ray, whereas classifier trained with \texttt{Locus} consistently sharpens attention onto the diagnostically relevant anatomical structures across both backbones. This highlights that the benefit of anatomy-guided regularization is most pronounced for architectures which more prone to background dominance, while still generalizing effectively across both convolutional and transformer-based architectures.

\begin{figure}[!ht]
    \centering
    \includegraphics[width=0.9\linewidth]{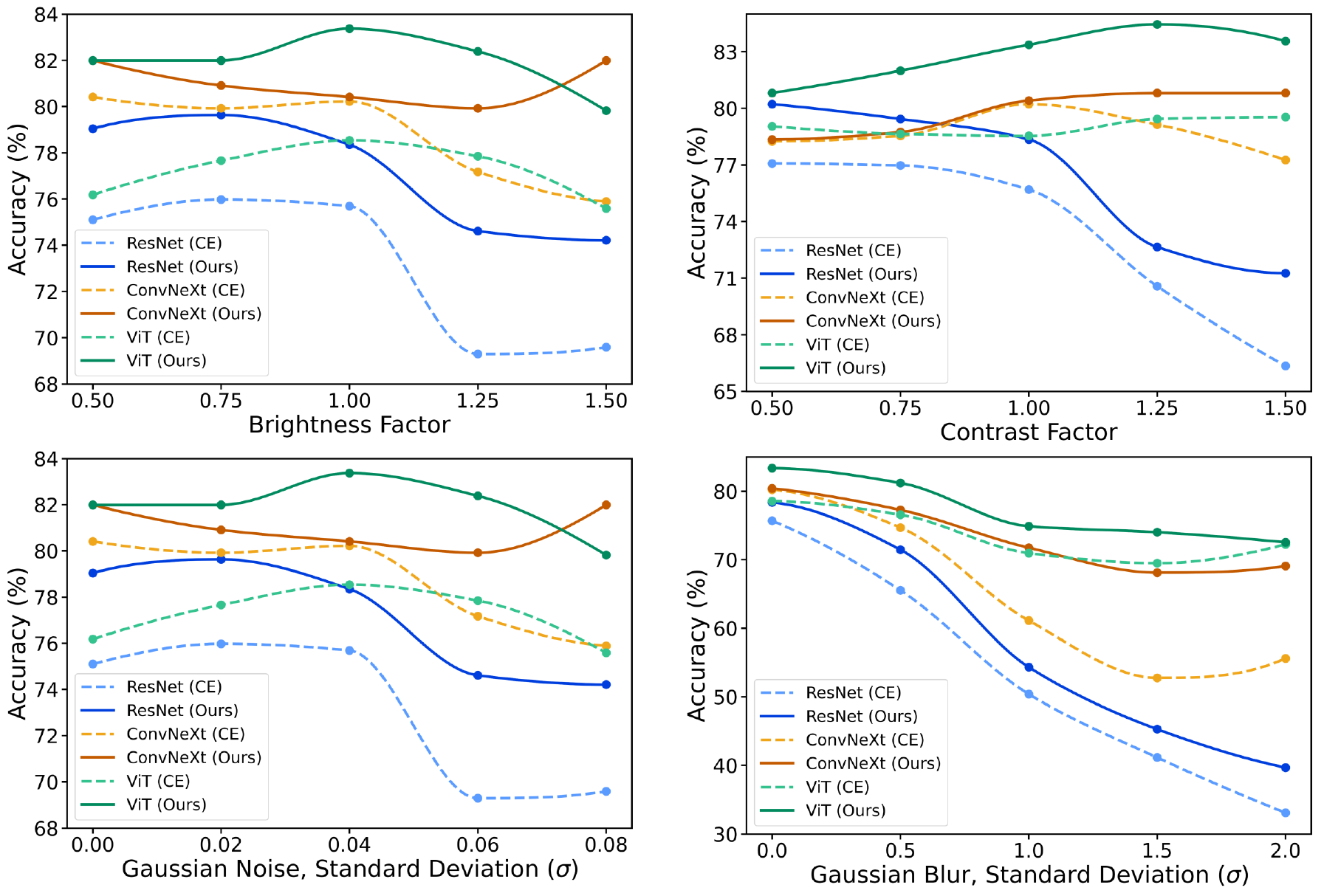}
    \caption{Robustness of our model on ChestX-ray under varying image perturbations: brightness (top left), contrast (top right), Gaussian blur (bottom left), and Gaussian noise (bottom right).}
    \label{fig:chexpert_robustness}
\end{figure}

Fig.~\ref{fig:chexpert_robustness} presents model robustness on Chest X-ray dataset under four types of image perturbations, including brightness, contrast, Gaussian blur, and Gaussian noise, applied only to the test set, across ResNet, ConvNeXt, and ViT backbones, comparing standard cross-entropy training (CE) against Ours. Across all perturbation types, classifiers trained with our proposed regularizer (Ours) consistently maintain higher accuracy as perturbation severity increases, while classifiers trained only with CE term (particularly ResNet) degrade sharply, most notably under Gaussian blur. This indicates that anatomy-guided attention regularization not only improves classification performance and interpretability but also enhances robustness to common imaging perturbations, likely by discouraging reliance on spurious or background-correlated features that are more sensitive to such distortions.

\begin{figure}[!ht]
    \centering
    \includegraphics[width=\linewidth, trim=0 0 0 0]{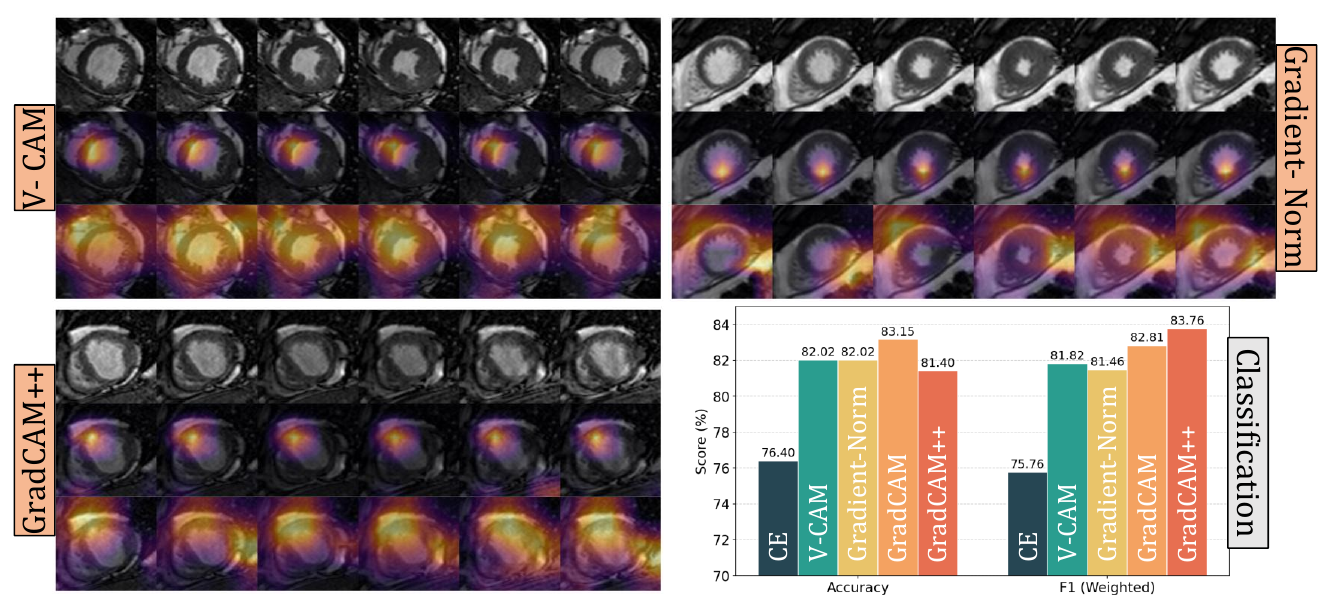}
    \caption{Generalizability of our approach across various attention extraction methods. At each block: input, CE, and Ours (top to bottom), shown for V-CAM, GradCAM++, and Gradient-Norm, with corresponding classification performance (bottom-right).}
    \label{fig:generalizability}
\end{figure}

Fig.~\ref{fig:generalizability} visualizes the generalizability of the proposed anatomy-guided regularizer across multiple attention extraction methods. Across all saliency techniques, including Vanilla-CAM (V-CAM), GradCAM, GradCAM++, and Gradient-Norm, our proposed approach consistently achieves superior classification performance, confirming that the proposed regularization is agnostic to the choice of attention map extraction method. Qualitatively, attention maps produced by our method (bottom row at each block) exhibit more focused and anatomically coherent activations compared to the diffuse, background-prone responses of CE (middle row), consistently across all saliency methods.

\begin{wraptable}{r}{0.45\textwidth}
\vspace{-20pt}
\centering
\caption{Effect of regularization strength ($\lambda$) on ResNet classification performance (\%) on BUSI dataset. }
\begin{tabular}{lccc}
\toprule
$\lambda$ & Accuracy & Precision & F1-score \\
\midrule
$0.1$ & $87.76$ & $88.18$ & $87.23$ \\
$0.3$ & $85.71$ & $87.14$ & $84.66$ \\
$0.5$ & $87.76$ & $88.18$ & $87.23$ \\
$0.7$ & $86.73$ & $87.28$ & $86.07$ \\
$0.9$ & $87.76$ & $88.18$ & $87.23$ \\
\midrule
CE only & $86.73$ & $86.86$ & $86.25$ \\
\bottomrule
\end{tabular}
\label{tab:lambda_ablation}
\vspace{-15pt}
\end{wraptable}

While the main results in Tab.~\ref{tab:acc_perf} use a single fixed $\lambda_{\max}=0.1$ across all datasets and backbones, Tab.~\ref{tab:lambda_ablation} studies sensitivity of \texttt{Locus} to the choice of $\lambda_{\max}$ specifically on the BUSI dataset~\cite{al2020dataset} under the ResNet backbone. At $\lambda = 0.1$, $0.5$, and $0.9$, our regularizer improves all three metrics over the CE-only baseline, peaking at 87.76\% accuracy. We observe performance decreases at $\lambda = 0.3$ and $0.7$, suggesting intermediate strengths can transiently over-constrain the classifier, though precision stays above the CE-only baseline throughout.

\begin{figure}[!ht]
    \centering
    \includegraphics[width=0.97\linewidth]{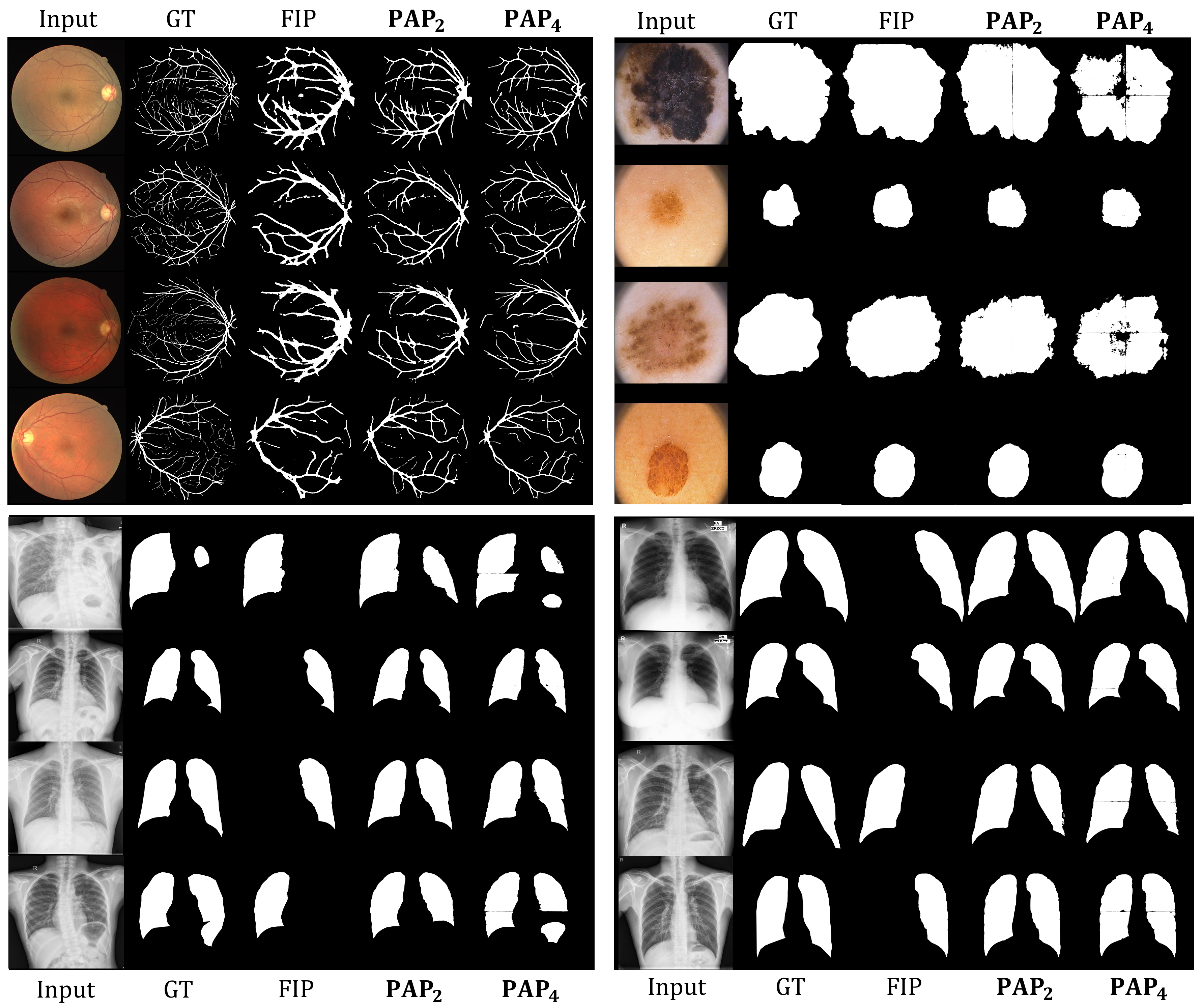}
    \caption{Qualitative comparison of Full Image Prompting (FIP) vs. Patch-wise Anatomical Prompting (PAP) on retinal vessel, skin lesion, and chest X-ray segmentation. PAP$_n$ denotes prompting with the image divided into $n$ patches.}
    \label{fig:patchAb}
\end{figure}

Restating our primary objective in Sec.~\ref{sec:PAP} of anatomical grounding rather than segmentation precision, Fig.~\ref{fig:patchAb} provides a comprehensive qualitative comparison of full image prompting (FIP) and our patch-wise anatomical prompting (PAP) across retinal, dermoscopic, and chest X-ray modalities. We utilize Medical SAM3~\cite{jiang2026medical} as the underlying SFM for prompting in all settings. We find that capturing fine-grained anatomical structure, such as retinal vasculature, benefits from a larger number of patches $n$, e.g., PAP$_4$ recovers substantially finer vessel branching than PAP$_2$, which in turn improves upon FIP, indicating that finer spatial decomposition helps the SFM attend to thin, distributed structures that are otherwise under-segmented when prompted on the full image.

Interestingly, we observe a failure mode specific to FIP in cases where the target anatomy consists of disjoint regions, as with the two lungs in chest X-ray images. Since FIP prompts the SFM over the entire image at once, we find that it frequently produces only one of the two lungs, rather than segmenting them separately. Our PAP formulation with $n>1$ largely resolves this, as constraining each prompt to a localized patch encourages the SFM to recover the two lungs as distinct, well-separated masks.

For the dermoscopic skin lesion images, PAP$_2$ performs comparably to FIP. However, we observe that when the internal texture of the lesion is heterogeneous and the classification signal depends primarily on the boundary contour, FIP outperforms PAP as partitioning the lesion across patches disrupts the SFM's view of the full boundary, leading to fragmented or over-segmented masks.

\begin{figure}[!ht]
    \centering
    \includegraphics[width=0.9\linewidth]{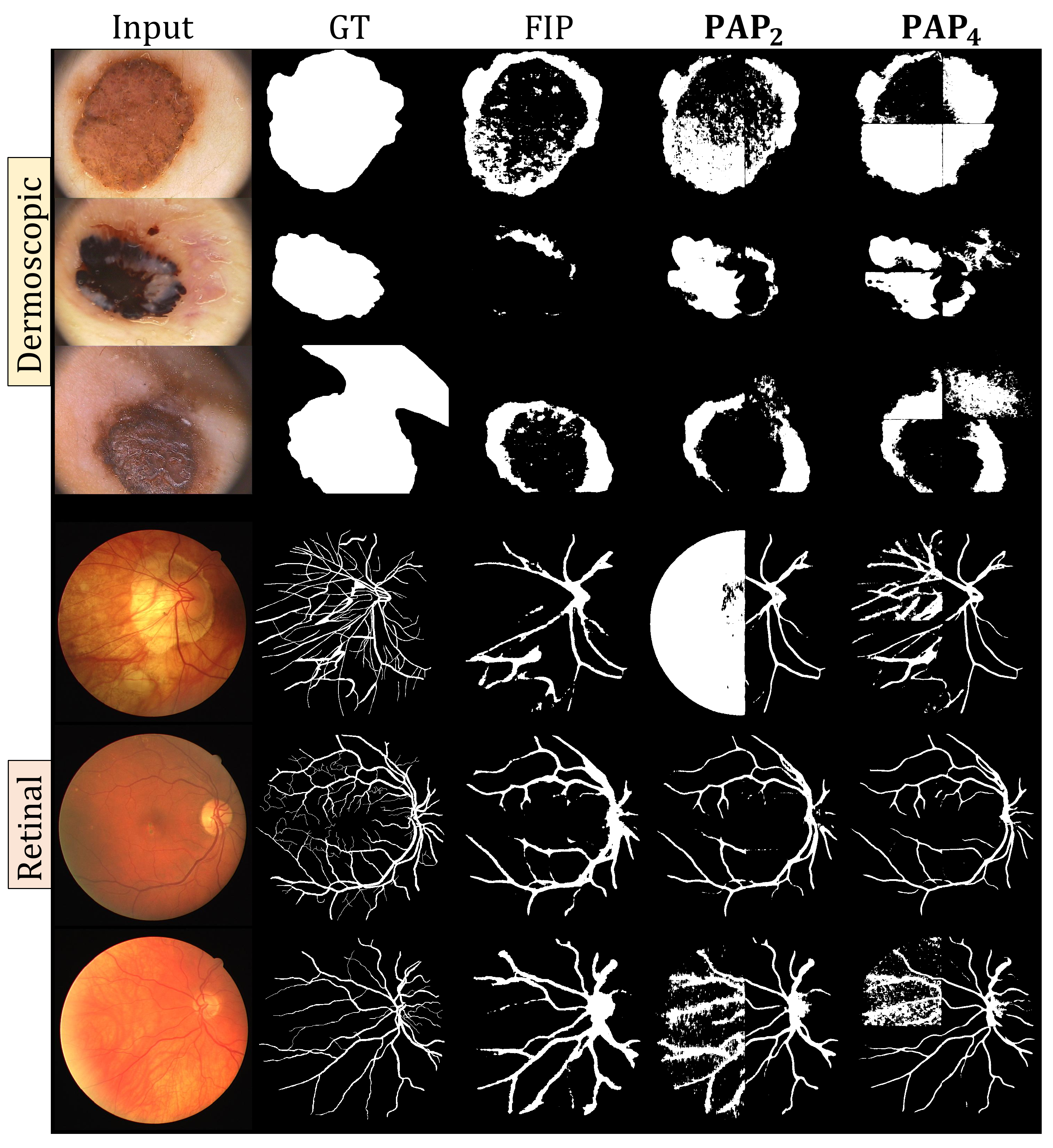}
    \caption{Failure cases of SFMs across dermoscopic and retinal images.}
    \label{fig:fc}
\end{figure}

We further examine representative failure cases in Fig.~\ref{fig:fc}. Both FIP and PAP tend to fail when the input exhibits low contrast between the ROI and surrounding tissue, highly heterogeneous internal texture (e.g., hair, ruler marks, or pigment variation in dermoscopic images), or ambiguous, low-contrast boundaries, causing the SFM to  poorly segment the region regardless of different prompting strategies. For PAP specifically, failures are compounded when the anatomical structure is fine and continuous (e.g., retinal vasculature), where the imposed patch grid can fragment otherwise connected structures along patch boundaries.

\section{Conclusion}
In this paper, we present \texttt{Locus}, an anatomy-guided attention regularization framework that leverages pretrained segmentation foundation models to steer medical image classifiers toward diagnostically relevant anatomical structures. Instead of enforcing rigid pixel-wise alignment, our hinge-based regularizer penalizes the classifier only when background attention dominates, while our patch-wise prompting strategy recovers spatially disjoint anatomical structures missed under full-image prompting. We validate \texttt{Locus} on eight diverse medical imaging datasets spanning dermoscopy, X-ray, histopathology, ultrasound, and cardiac MRI, showing consistent improvements in classification performance and attention interpretability over SOTA baselines. Moreover, these improvements are consistent across both convolution/transformer-based backbones, saliency methods, and test-time perturbations.\\

\noindent \textbf{Limitations.} \texttt{Locus} relies on a text prompt to the SFM to define the anatomically relevant foreground, which presumes a single, well-defined target region exists for a given task. This assumption does not always hold, and we identify two related limitations: (i) for some applications, what counts as relevant is itself ambiguous, e.g., in tumor microscopy, it is unclear whether only the tumor tissue, or also the surrounding microenvironment, should be treated as foreground; since $M$ is derived directly from the chosen prompt, an inappropriate or overly narrow prompt can misdirect $\mathcal{R}$ toward the wrong region; and (ii) our framework presumes the classification label is correlated with a specific anatomical structure or shape, an assumption that fails when the class distinction is not tied to any particular anatomical region, in which case the benefit of anatomical guidance is unclear.

Future work includes (i) exploring more principled uncertainty-aware weighting to handle cases where anatomical guidance is limited by the SFM's generalization; (ii) jointly learning the number of patches in our prompting strategy, rather than selecting it empirically; (iii) incorporating other text-guided segmentation foundation models for tasks beyond classification; and (iv) extending \texttt{Locus} to settings where the classification task is not anatomy-guided but instead depends on subtle or hidden cues that cannot be readily specified as an anatomical prompt to the SFM.\\

\noindent \textbf{Acknowledgments.} This work was supported by NSF CAREER Grant $2239977$ and NIH 1R21EB032597.\\

\noindent \textbf{Disclosure of Interests.} The authors have no competing interests to declare that are relevant to the content of this article.

\bibliographystyle{splncs04}
\bibliography{references}
\end{document}